\documentclass[letterpaper, 10 pt, conference]{ieeeconf}
\IEEEoverridecommandlockouts 
\usepackage[utf8]{inputenc}

\usepackage{comment}
\usepackage{xspace}
\usepackage{graphicx}
\usepackage{overpic}
\usepackage{svg}
\usepackage{xcolor}
\usepackage{bm}
\usepackage[hang,flushmargin]{footmisc}
\usepackage{booktabs}
\usepackage{tabularx}

\usepackage{tabularx}
\usepackage{multirow}
\usepackage{booktabs}
\usepackage{colortbl}
\usepackage{float}
\usepackage{placeins}

\usepackage{array}
\newcolumntype{P}[1]{>{\centering\arraybackslash}p{#1}}  

\usepackage{etoolbox}
\makeatletter
\patchcmd{\@makecaption}
{\scshape}
{}
{}
{}
\newcommand{\onetagright}{\tagsleft@false}
\makeatother

\usepackage{amsmath}
\usepackage{amssymb}
\usepackage{amsthm}
\usepackage{siunitx}

\usepackage{algorithm}
\usepackage{algpseudocode}

\usepackage[
	activate   = {true},
	protrusion = false,
	expansion  = true,
	kerning    = true,
	spacing    = true,
	tracking   = false,
	auto       = true,
	selected   = true,
	factor     = 2000,
	stretch    = 50,
	shrink     = 20,
]{microtype}

\definecolor{gg}{RGB}{240, 74, 0}

\makeatletter
\let\NAT@parse\undefined
\makeatother
\usepackage[pdfa,colorlinks,bookmarksopen,bookmarksnumbered,allcolors=gg]{hyperref}
\usepackage[english]{babel}

\usepackage[nameinlink,capitalise]{cleveref}
\crefname{line}{line}{lines}
\crefname{figure}{Fig.}{Figs.}
\Crefname{figure}{Fig.}{Figs.}
\crefname{equation}{Eq.}{Eqs.}
\Crefname{equation}{Eq.}{Eqs.}
\crefname{section}{Sec.}{Secs.}
\Crefname{section}{Sec.}{Secs.}
\crefname{definition}{Def.}{Defs.}
\Crefname{definition}{Def.}{Defs.}
\crefname{algorithm}{Alg.}{Algs.}
\Crefname{algorithm}{Alg.}{Algs.}
\crefname{assumption}{Asm.}{Asms.}
\Crefname{assumption}{Asm.}{Asms.}
\crefname{subassumption}{Asm.}{Asms.}
\Crefname{subassumption}{Asm.}{Asms.}
\Crefname{problem}{Problem}{Problems}
\crefname{problem}{Problem}{Problems}

\usepackage{flushend}

\let\labelindent\relax
\usepackage[inline]{enumitem}
\usepackage{mathtools}

\usepackage{arydshln} 

\title{\fontsize{17pt}{24pt}\selectfont \bf
RAG-Modulo: Solving Sequential Tasks \\[-0.2em]
using  Experience, Critics, and Language Models
}

\author{Abhinav Jain, Chris Jermaine$^*$, Vaibhav Unhelkar$^*$
\thanks{*Equal advising, authors listed alphabetically.}
\thanks{Corresponding author: {\tt\small abhinav.jain@rice.edu}}
\thanks{All authors are affiliated with the Department of Computer Science, Rice University, Houston, TX. This work was supported in part by the NSF and Rice University funds.}
\thanks{This work has been submitted to the IEEE for possible publication. Copyright may be transferred without notice, after which this version may no longer be accessible.}}

\begin{document}
\maketitle
\thispagestyle{empty}
\pagestyle{empty}

\newcommand{\approach}{RAG-Modulo\xspace}

\begin{abstract}
Large language models (LLMs) have recently emerged as promising tools for solving challenging robotic tasks, even in the presence of action and observation uncertainties. Recent LLM-based decision-making methods (also referred to as LLM-based agents), when paired with appropriate critics, have demonstrated potential in solving complex, long-horizon tasks with relatively few interactions. However, most existing LLM-based agents lack the ability to retain and learn from past interactions—an essential trait of learning-based robotic systems. 
We propose \approach, a framework that enhances LLM-based agents with a memory of past interactions and incorporates critics to evaluate the agents' decisions. The memory component allows the agent to automatically retrieve and incorporate relevant past experiences as in-context examples, providing context-aware feedback for more informed decision-making. Further by updating its memory, the agent improves its performance over time, thereby exhibiting learning. Through experiments in the challenging BabyAI and AlfWorld domains, we demonstrate significant improvements in task success rates and efficiency, showing that the proposed \approach framework outperforms state-of-the-art baselines.
\end{abstract}

\section{Introduction}

Solving goal-driven sequential tasks is a core problem in robotics, with a wide array of challenges~\cite{chevalier2018babyai, shridhar2020alfworld, shridhar2020alfred, zhu2020robosuite, jiang2023vima, heo2023furniturebench}. Due to imperfect actuation, real-world robots operate in stochastic environments. Their sensors often provide only a partial view of the surroundings, requiring decision-making under partial observability and limited knowledge of the world model. To reduce the programming burden for end-users, even complex, long-horizon tasks are frequently defined by sparse reward functions or natural language descriptions of the robot's goal.


%
Various paradigms and corresponding methods have been explored to address this fundamental challenge~\cite{argall2009survey, kober2013reinforcement, ravichandar2020recent, garrett2021integrated, singh2022reinforcement}. The planning paradigm assumes access to a task model, which is often unavailable in real-world applications. While reinforcement learning can operate without a task model, it typically requires a prohibitively large number of exploratory interactions and significant manual effort for reward design. This challenge is further compounded in partially observable environments, where sparse rewards and safety concerns limit the feasibility of extensive exploration.

To complement these long-standing paradigms, language models have recently emerged as promising tools for solving long-horizon tasks in robotics \cite{yao2022react, ahn2022can, huang2022language, carta2023grounding, singh2023progprompt, song2023llm, lin2023text2motion, vemprala2024chatgpt}. They can approximate world knowledge \cite{petroni2019language, roberts2020much, jiang2020can} and use few-shot reasoning to decompose high-level tasks into mid-level plans \cite{kojima2022large, wei2022chain, zhou2022least}. Additionally, they can function as dynamic planners, adjusting their strategies based on environmental feedback, which is especially useful in partially observable settings \cite{song2023llm}. Moreover, their performance is shown to improve when integrated with formal systems that evaluate decisions based on criteria such as correctness, executability, and user preferences \cite{kambhampati2024llms}.

\begin{figure}[t]
  \centering
  \includegraphics[scale=0.27]{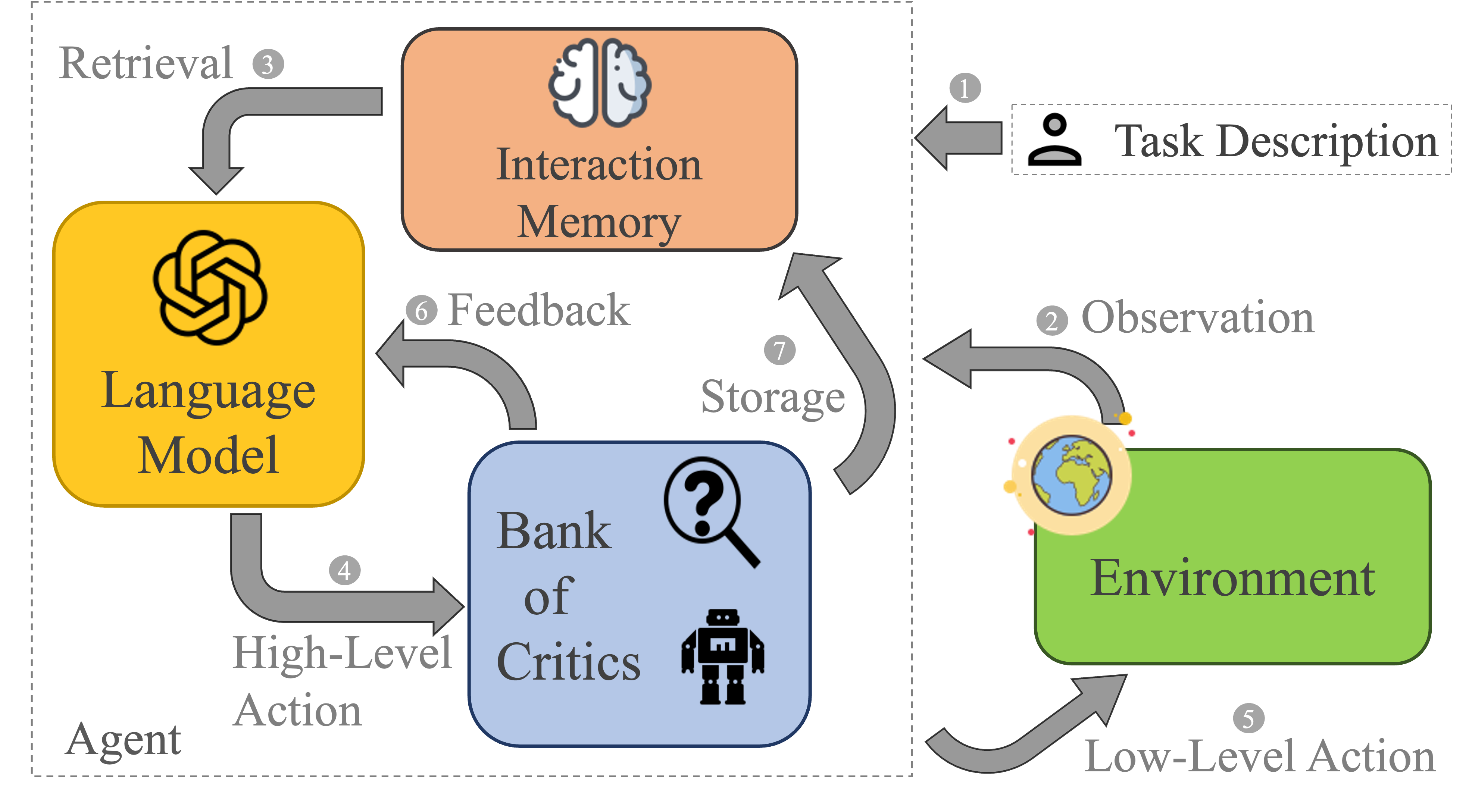}
  \caption{The \approach framework incorporates a language model to generate candidate actions and a set of critics to evaluate them. Importantly, it features mechanisms for storing and retrieving past interactions, which enable learning from experience and improve decision-making over time.} 
  \label{fig:framework}
\end{figure}

Despite their promise, most existing LLM-based decision-making methods (also referred to as LLM-based agents) lack the ability to learn from experience.
To effectively solve complex, long-horizon tasks, a robotic agent must demonstrate the ability to learn: meaning it should improve its performance over time as it gains more experience in its environment.
A prevalent approach to realize such ``learning'' for LLM-based robotic agents is to tune prompts using in-context examples \cite{arenas2024prompt}, but this method is constrained by the selection of examples, requires domain knowledge, and demands manual effort.
Another option is to fine-tune language models based on past interactions \cite{carta2023grounding, pallagani2022plansformer}, but this approach can be computationally expensive and resource intensive.
To address these gaps, we propose \approach: a framework which augments a language model with a memory that stores past interactions, retrieving relevant experience at each step of the task to guide robot decision-making.

\begin{figure*}[ht]
  \centering
    \begin{minipage}[b]{0.60\textwidth}
    \centering
    \includegraphics[width=\textwidth]{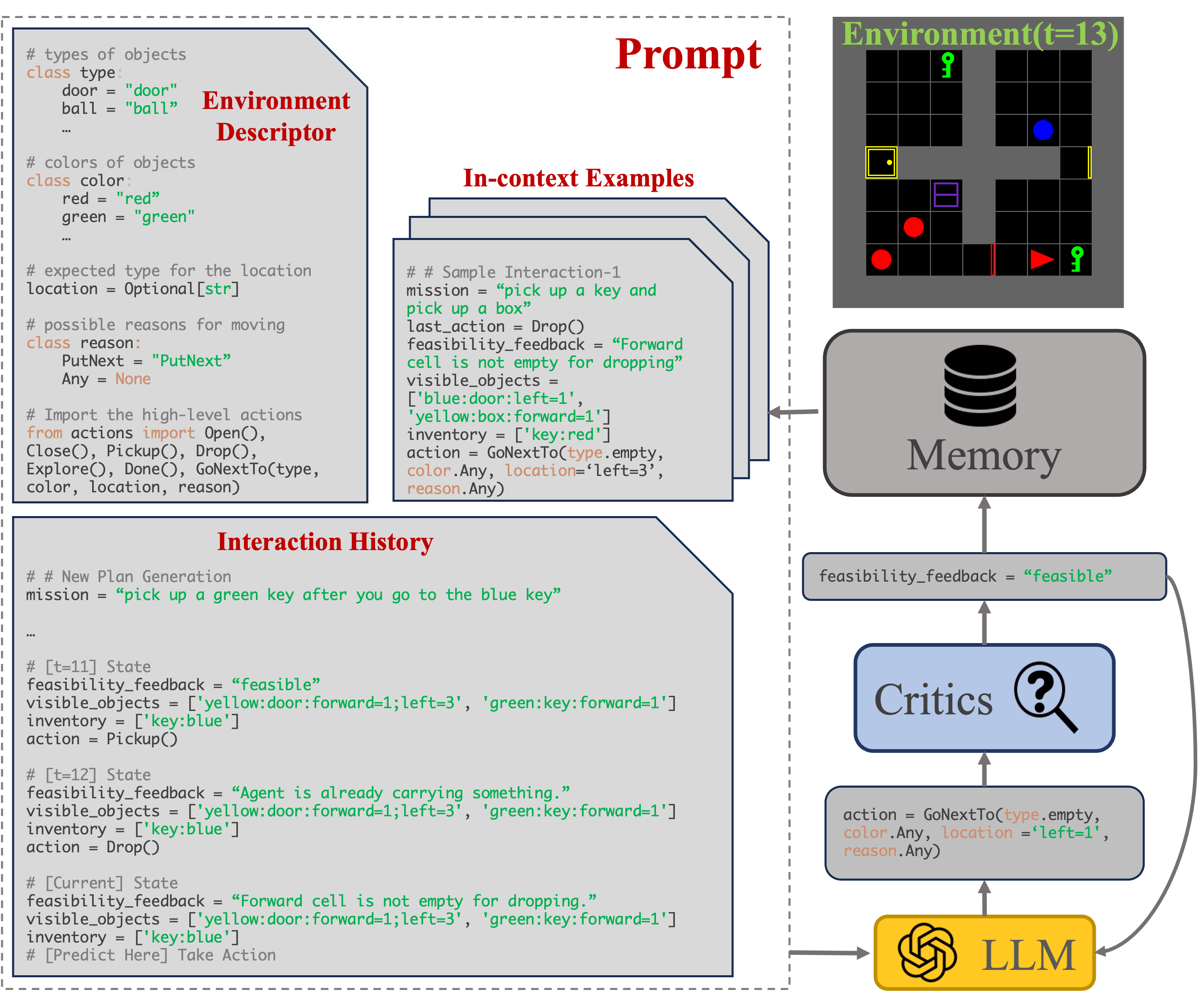}
  \end{minipage}
  \hspace{0.05\textwidth} 
  \begin{minipage}[b]{0.30\textwidth}
    \centering
    \includegraphics[width=\textwidth]{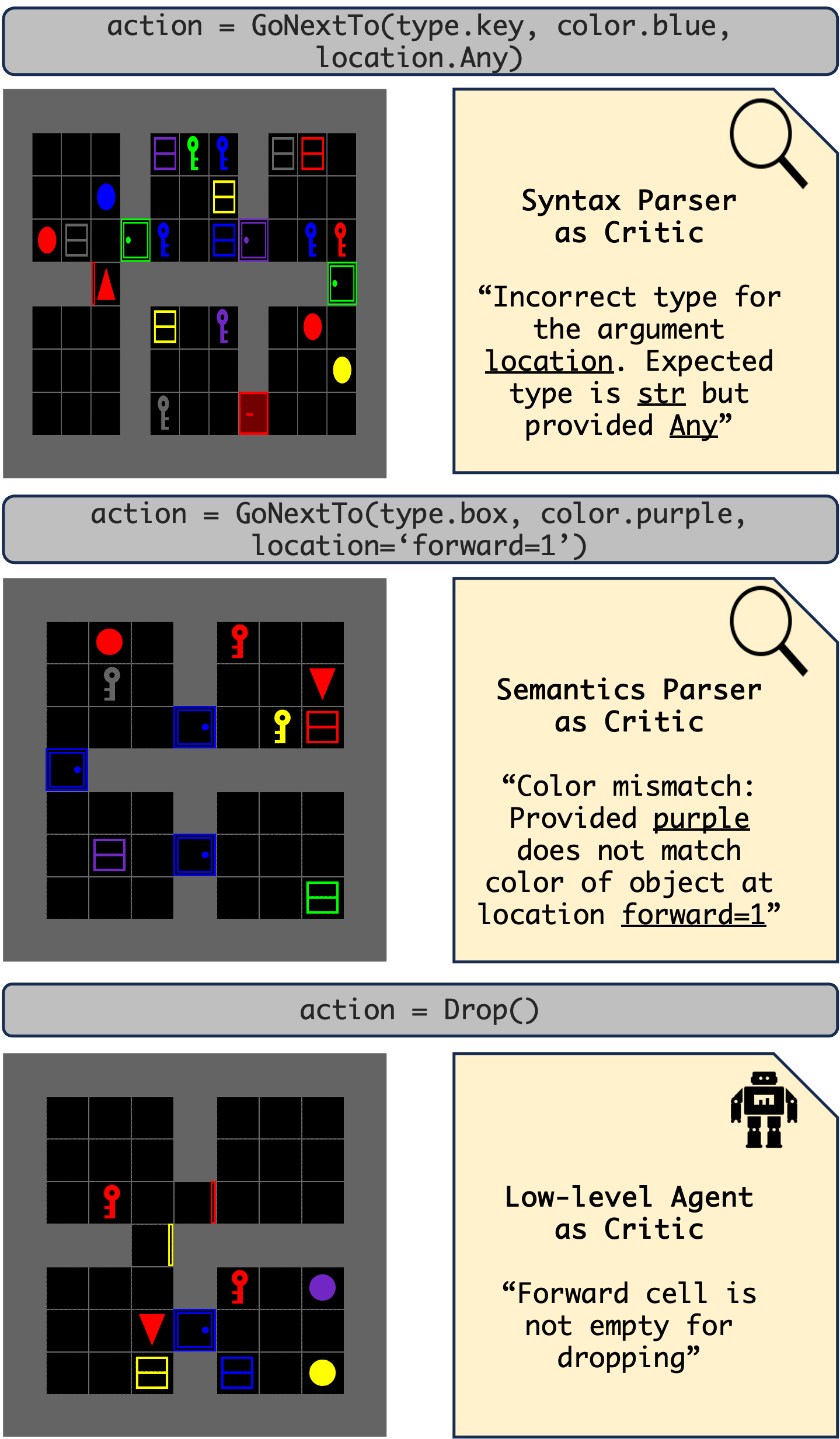}
  \end{minipage} 
  \caption{(Left) The prompt in \approach consists of an environment descriptor, a history of past interactions, and in-context examples to guide the LLM in selecting a feasible action. Here, the agent can be carrying a blue key, which it needs to drop before picking up the green key. The retrieved in-context example shows a similar scenario where the agent is unable to drop an object in an occupied cell. Based on this, the agent generates an action to move to an empty cell before completing the task. (Right) Illustration of how each critic provides feedback for the infeasible action shown on top. }
  \label{fig:illustration}
\end{figure*}

As shown in \cref{fig:framework,fig:illustration}, \approach extends the LLM-Modulo framework \cite{kambhampati2024llms} with memory, where formal verifiers or critics evaluate the feasibility of actions at each step based on criteria like syntax, semantics, and executability. The interactions, along with feasibility feedback, are stored in memory and retrieved as in-context examples, enabling automatic prompt tuning for future tasks. By leveraging these past interactions, the agent can generalize from its experiences, avoid repeated mistakes, and make more accurate decisions—much like how humans learn from their past errors.
In summary, building on the insight of memory-augmented behavior generation, this paper makes three key contributions: 
\begin{itemize} 
    \item \approach: A framework with LLM-based agents that learns not through back-propagation, but by building up a database of experiences (Interaction Memory) that it then accesses.
    \item A retrieval mechanism that enable LLM-based agents to access context-aware interactions from memory as in-context examples, automatically tuning prompts and reducing manual effort. 
    \item A suite of experiments on challenging tasks from AlfWorld and BabyAI, where \approach outperforms recent baselines and demonstrates improved performance with minimal environment interactions. 
\end{itemize}

\section{Problem Formulation}

In this section, we formally model the tasks of interest and define the problem, followed by an explanation of how language models can be prompted to function as agents.

\subsection{Task Model}
We focus on object-centric, goal-driven sequential robotic tasks that may involve uncertainties in both actions and observations~\cite{mandlekar2023mimicgen}.
More specifically, we denote $\mathbb{S}_{o}$ as the set of all possible objects in the robot's environment and $\mathbb{S}_{p}$ as the set of object properties.
We formally define the task model with the tuple $(\mathbb{S}, \mathbb{G}, \mathbb{A}, \mathbb{O}, \mathrm{T}, \mathrm{R}_g, h, \gamma)$.
Given $\mathbb{S}_{o}$ and $\mathbb{S}_{p}$, a state $s\in\mathbb{S}$ is defined as an assignment of object properties.
$\mathbb{A}$ is the set of low-level physical actions and $\mathbb{G}$ is the set of all goals. A goal $g\in \mathbb{G}$ is the natural language description of the goal state. $\mathbb{O}$ is the set of observations retrieved from states via an observation function $\mathrm{O}: (\mathbb{S} \times \mathbb{A}) \mapsto \mathbb{O}$, and $\mathrm{T}: (\mathbb{S} \times \mathbb{A}) \mapsto \mathbb{S}$ is the transition function. $\mathrm{R}_g$ is the goal-conditioned reward function, which $=1$ when goal is achieved, else $0$.
Finally, $\gamma$ denotes the discount factor and $h$ represents the task horizon.


Following prior work \cite{shridhar2020alfworld, singh2023progprompt}, the agent is also equipped with a set of high-level text actions, denoted by $\mathbb{C}$. In reinforcement learning (RL) literature, these can be interpreted as macro actions or options \cite{sutton1999between, daniel2016probabilistic}. Each action $c \in \mathbb{C}$ is composed of a function and its corresponding set of arguments, i.e., $c = \textsc{function(argument)}$, such as $\textsc{Open(type.door, color.red)}$. We assume that the robot can execute this high-level action by breaking it down into a sequence of primitive actions $(a_1, a_2, \ldots)$, governed by its low-level policy $\pi_c$, until a termination condition $\beta_c$ is met. For the remainder of the paper, we simply refer to high-level actions as actions.

\subsection{Problem Statement}
We can now formally define the problem statement. Given the initial state, $s_0$, generate the shortest sequence of actions $(c_1, c_2, \ldots, c_t)$ to reach the goal state described as $g$. 



\subsection{Language Models as Agents} 
As shown in recent works~\cite{singh2023progprompt, song2023llm}, large language models $(LLM)$ can be prompted at each time step to generate a sequence of actions using the following prompt:  
\begin{align*}
    \textsc{prompt}_t = p_{env}; \{g^k, o^k, c^k\}_{k=1}^K; g; \{o_{1:t-1}, c_{1:t-1}\};o_t
\end{align*}
where, at time-step $t$, the prompt consists of (i) fixed prefix $p_{env}$ describing the environment; (ii) $K$ in-context examples comprised of goal-observation-action tuples, $\{g^k, o^k, c^k\}_{k=1}^K$; (iii) goal description $g$; (iv) the history of actions for previously visited states, $\{o_{1:t-1}, c_{1:t-1}\}$; (v) the current observation, $o_t$. The in-context examples demonstrate how to solve similar tasks and the history of past interactions provides the language model with context about how the agent has interacted with the environment so far. 
\section{Related Work}


In this section, we discuss related methods that utilize language models as agents, use memory components and incorporate retrieval-augmented generation (RAG).

\textbf{Language Models as Agents.} Recent works have explored using language models as agents for solving long-horizon tasks by generating plans ~\cite{huang2022language, singh2023progprompt, song2023llm, vemprala2024chatgpt, lin2023text2motion}. Approaches like ProgPrompt \cite{singh2023progprompt, hazra2024saycanpay} generate static plans offline, which may fail when encountering unforeseen object interactions in a partially observable environment. LLM-Planner-like approaches \cite{song2023llm, vemprala2024chatgpt, lin2023text2motion} offer a more online approach, allowing for plan updates if an action fails, but it does not store past successes and failures to guide future decisions. The method in \cite{vemprala2024chatgpt} involves a human in the loop to prompt and verify. \cite{lin2023text2motion} generates feasible plans but relies on precise model dynamics estimation to assess plan feasibility. More recently, \cite{kambhampati2024llms} have shown that language models should be coupled with verifiers or critics to generate sound plans. These recent methods have informed our work; however, in contrast to these works, \approach stores and retrieves past interactions from memory to inform and improve decision-making.




\textbf{Learning with Experience.}  Reinforcement learning agents typically use a replay buffer to store experiences for policy optimization. However, solving complex long-horizon tasks often demands millions of trajectories or environment interactions to learn effectively \cite{chevalier2018babyai}. In contrast, our approach requires only a few hundred experiences to enable meaningful learning. Very recently, some LLM-based approaches have introduced memory modules that store past experiences and expand as the agent interacts with the environment \cite{zhang2023bootstrap, wang2023voyager, tziafas2024lifelong, shinn2024reflexion, zhao2024expel}. These methods store experiences at the skill level, retrieving them when needed, but lack the ability to track past successes and failures at the interaction level. Moreover, they often require multiple LLMs to reason, relabel and abstract primitive skills into more complex composite ones.

In contrast, the proposed \approach stores experiences at the interaction level, removing the need for LLM-guided relabeling, and retrieves these experiences at every decision-making step to offer more informed guidance to the language model. Importantly, our work is complementary to these methods, as they tackle different aspects of continual learning --- one focuses on learning library of skills, while the other emphasizes learning from past mistakes and successes.

\textbf{RAG systems for Robotics.} Retrieval Augmented Generation (RAG) systems enhance language model predictions by retrieving relevant information from external databases \cite{izacard2023atlas, borgeaud2022improving}. For example, \cite{kim2024rada} employs RAG to collect exemplars for solving sub-tasks with web agents, while \cite{dai2024vistarag} retrieves driving experiences from a database for autonomous vehicle planning. In robotics, \cite{goyal2022retrieval} explores retrieval for deep RL agents, but it does not use LLMs, limiting its adaptability and scalability. \cite{zhu2024retrieval} employs a policy retriever to extract robotic policies from a large-scale policy memory.
In contrast, our approach integrates a RAG system within an LLM-Modulo framework, where past interactions and feedback from critics is stored and continuously expanded. This enables the retrieval of interaction-level experiences, including mistakes and corrections, providing more detailed and context-aware guidance for sequential decision-making.

\begin{figure*}[htbp]
  \centering
    \begin{minipage}[b]{0.20\textwidth}
    \centering
    \includegraphics[width=\textwidth]{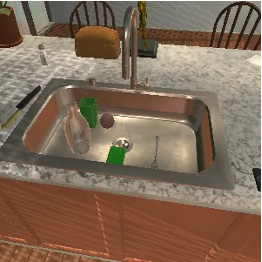}
  \end{minipage}
  \hspace{0.01\textwidth} 
  \begin{minipage}[b]{0.75\textwidth}
    \centering
    \includegraphics[width=\textwidth]{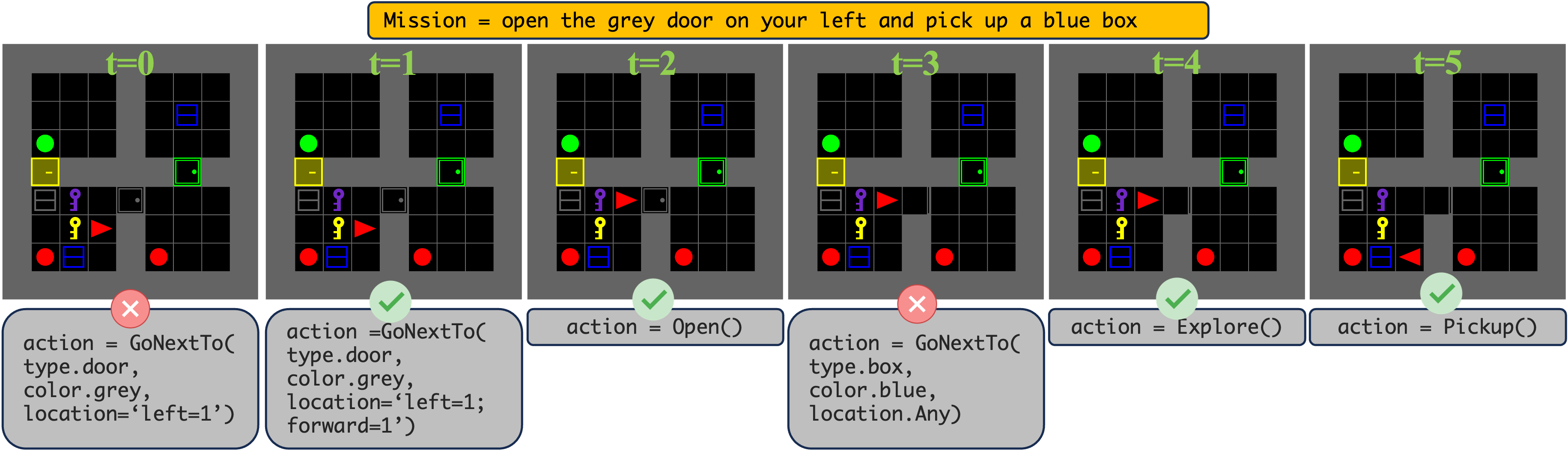}
  \end{minipage}
  \caption{(Left) AlfWorld Domain where the agent is shown in a household environment. (Right) Execution trace while solving a task from BabyAI. Ticks and Crosses show feasible and infeasible actions respectively.}
  \label{fig:domains} 
\end{figure*}

\section{Proposed Approach}

\algnewcommand{\Try}{\textbf{try:}}
\algnewcommand{\Except}{\textbf{except}}
\begin{algorithm}
\caption{\approach}
\begin{algorithmic}[1]
\State INPUT: $(g, h, LLM, \mathbb{M})$
\State $t\gets 1$ \Comment{Initialize the time-step}
\State $M_{}\gets\{\}$ \hfill 
\While{$t\leq h$ or ($g$ is satisfied)}
    \State $o_t \gets$ Observe the environment
    \State $(I^{k}, c^{k})\gets$\hspace{-0.1em} Retrieve interactions from memory (Eq.~\ref{eq:retrieval})
    \State $\textsc{prompt}_t \gets$ Construct the prompt (Eq.~\ref{eq:prompt})
    \State $c_t \gets LLM(\textsc{prompt}_t)$ \Comment{Predict action}
    \State $f_{t} \gets $ \textsc{CheckFeasibility} $(o_t, c_t)$
    \If{$f_{t}$ is SUCCESS} \Comment{Keep track of interactions}
        \State $M_{} \gets M_{} \cup \{I_{t}\doteq(g,c_{t-1}, f_{t-1}, o_t), c_t\}$ 
    \EndIf
\EndWhile
\If{$g$ is satisfied}
    \State $\mathbb{M} \gets M_{}\cup \mathbb{M}$ \Comment{Update memory}
\EndIf \State
\Return $(c_{1:t}, \mathbb{M})$
\end{algorithmic} 
\label{alg:generate}
\end{algorithm}

\begin{algorithm}
\caption{\sc CheckFeasibility}
\begin{algorithmic}[1]
\State INPUT: $o_t, c_t$
\State $f_{t} \gets \textsc{success}, \textsc{reason} \gets \textsc{None}$
\State \textbf{try:} \hfill
  \State {\quad Parse $c_t$ using $\varphi_{syntax}$} \Comment{Syntax Critic}
  \State {\quad Parse $c_t$ using $\varphi_{semantics}$} \Comment{Semantics Critic}
  \State {\quad \textbf{repeat}}
    \State {\qquad Execute $\pi_{c_t}$} \Comment{Low-level Policy Critic}
  \State{\quad \textbf{until} $\beta_{c_t}$ is $\mathrm{True}$}
\State \Except \;Exception as \textsc{reason:}
    \State \quad $f_{t} \gets \textsc{failure(reason)}$    \State
\Return $f_t$
\end{algorithmic} 
\label{alg:critique}
\end{algorithm}

We now describe \approach, summarized in \cref{alg:generate}, which is composed of an LLM, a bank of critics, and an interaction memory $(\mathbb{M})$ coupled with mechanisms for storing and retrieving interaction experience.
At each step $t$ of a task specified by natural language goal $g$ and horizon $h$, \approach first retrieves interactions $I$ from the memory that are relevant to the  task and current observation $o_t$, using them to guide the LLM's decision-making (line $6$ in \cref{alg:generate}). The LLM selects action $c_t$ based on this context and receives feedback (lines $8-9$) from a bank of critics (\cref{alg:critique}). If feasible, the interaction is stored (lines $10-12$). Once the goal is achieved, the interaction memory is updated for future retrieval (lines $14-16$), enabling learning from experience.

\subsection{Critics and Feedback}
Informed by the \cite{kambhampati2024llms}, \approach includes a bank of critics $(\varphi_{syntax}, \varphi_{semantics}, \varphi_{low-level})$ who provide feedback on actions selected by the LLM. $\mathbb{F}$ denotes the set of feedbacks described in natural language. The syntax parser $\varphi_{syntax}:\mathbb{C}\mapsto \mathbb{F}$ returns feedback based on syntactical correctness. It ensures that the LLM’s response adheres to the grammar rules of the environment. The semantics parser $\varphi_{semantics}:(\mathbb{C}\times\mathbb{O})\mapsto \mathbb{F}$ returns feedback based on semantic correctness. It verifies that the predicted action is meaningful and logically consistent with the current observation $o$, e.g., ensuring the agent has the correct key before opening a door. The low-level policy critic $\varphi_{low-level}:(\mathbb{C}\times\mathbb{O})\mapsto \mathbb{F}$ checks if $c$ is executable from $o$. It runs the execution using $\pi_c(o)$ until $\beta_c(o)$ is satisfied. For example, while traversing a path it can determine if an obstacle is encountered. 
As summarized in \cref{alg:critique}, each critic $\varphi$, either returns $\textsc{success}$ or \textsc{failure} along with the corresponding \textsc{reason}. This mimics how programmers receive feedback
from compilers during debugging. We now formally define the overall feasibility feedback $f\in\mathbb{F}$ as a function of the feedback from all critics:
\begin{equation*}
    f=
    \begin{cases}
      \textsc{success}, \qquad \text{if}\; \varphi_{syntax} \wedge \varphi_{semantics} \wedge \varphi_{low-level} \\
      \textsc{failure(reason)}, \text{otherwise}
    \end{cases}
\end{equation*}
Given the feedback, the prompt has the following structure:
\begin{align} \label{eq:prompt}
&\textsc{prompt}_t =  \\
 &p_{env};\{g^k, {c}_{-1}^k, f^k, o^k, c^k\}_{k=1}^K; g; (o_{1:t-1}, c_{1:t-1}, f_{1:t-1}); o_t \nonumber
\end{align}
where,  in-context examples and history now include the previous action $c_{-1}$ and its feasibility feedback. 

\subsection{Interaction Memory, Storage and Retrieval}
\approach considers a database of past interactions representing the agent’s memory $\mathbb{M}$ of solving prior tasks and their outcomes. We represent such interaction with the tuple $(I, c)$, where $I = (g, c_{-1}, f, o)$. Formally, the memory includes a set of interactions $\mathbb{M} = \{(I^1, c^1), \ldots, (I^m, c^m)\}$, where $m$ represents the memory size. 

\textbf{Retrieval.}  At every decision-making step of a given task, \approach retrieves from the memory the top-$K$ most relevant interactions $\{I^k, c^k\}_{k=1}^K $ that resemble the current task and situation and uses them as in-context examples as shown in \cref{fig:illustration}. Formally, this is represented as:
\begin{equation}
    I_{1:K} = \underset{I \in \mathbb{M}}{\operatorname{argmax_K}} \, \cos(e(I_t), e(I))
    \label{eq:retrieval}
\end{equation}
\noindent $\operatorname{argmax_K}$ returns the top-$K$ samples from the memory that have the highest cosine similarity with $I_t$. $e(I)$ represents the fixed-size embedding of $I$ generated by the encoder model $e$. As detailed in \cref{sec:experiments}, we use OpenAI's \textsc{text-embedding-3-large}  \cite{text-embedder} as the encoder model $e$ for realizing \approach in our experiments.


\textbf{Storage.} For every successfully completed task, $\{g, c_{1:h}, f_{1:h}, o_{1:h}, c_{1:h})\}$, \approach fills the memory with its interactions $(I_t, c_t)$ for which the current option $c_t$ is always feasible (i.e. $f(c_t)=\textsc{success}$). Thus, every stored tuple is a successful interaction that includes rectifications when $f_{t-1}=\textsc{failure}$, which can be used by the LLM when planning future actions.


\section{Experimental Setup}
\label{sec:experiments}

\begin{table*}
\centering
    \begin{tabular}{lcccccc}
    \\
    \toprule
     Approach &  \multicolumn{3}{c}{BabyAI-Synth}  &  \multicolumn{3}{c}{BabyAI-BossLevel}\\
     \cmidrule(lr){2-4}  \cmidrule(lr){5-7}
     & SR $\uparrow$ & InExec $\downarrow$ & Len $\downarrow$ & SR $\uparrow$ & InExec $\downarrow$ & Len $\downarrow$\\
    \midrule
    Expert & 0.96 & 0.79 & 8.34 & 0.98 & 0.43 & 8.41\\
    \hdashline
    ProgPrompt & $0.24 \pm 0.08$ & $-$ & $-$ & $0.11\pm 0.06$ & $-$ & $-$ \\
    LLM-Planner  & \textbf{0.48} $\pm$ \textbf{0.1} & $12.02\pm2.17$ & \textbf{14.72} $\pm$ \textbf{2.13} & $0.24\pm 0.08$ & $13.98\pm 1.48$ & $16.16\pm 1.35$ \\
    Ours & \textbf{0.48} $\pm$ \textbf{0.1} & \textbf{5.18}$\pm$ \textbf{1.18} & $14.82\pm 2.14$ & \textbf{0.57} $\pm$ \textbf{0.10} & \textbf{3.74} $\pm$ \textbf{0.78}& \textbf{12.48} $\pm$ \textbf{1.49}\\ \midrule\midrule
     & \multicolumn{3}{c}{AlfWorld-Seen} & \multicolumn{3}{c}{AlfWorld-Unseen} \\
     \cmidrule(lr){2-4}  \cmidrule(lr){5-7} 
     & SR $\uparrow$ & InExec $\downarrow$ & Len $\downarrow$ & SR $\uparrow$ & InExec $\downarrow$ & Len $\downarrow$\\
    \midrule
    Expert & 0.82 & 0.16 & 18.18 & 0.81 & 0.28 & 18.47 \\
    \hdashline
    ProgPrompt & $0.09\pm 0.05$ & $-$ & $-$ & $0.08\pm 0.05$ & $-$ & $-$ \\
    LLM-Planner  & $0.2\pm 0.07$ & $21.24\pm 1.76$ & $25.65\pm 1.46$ & $0.17\pm 0.06$ & $21.73\pm 1.7$ & $26.36\pm 1.4$ \\
    Ours & \textbf{0.52} $\pm$ \textbf{0.08} & \textbf{5.36} $\pm$ \textbf{1.39} & \textbf{20.54} $\pm$ \textbf{1.71} & \textbf{0.54} $\pm$ \textbf{0.09} & \textbf{7.17} $\pm$  \textbf{1.73} & \textbf{19.64} $\pm$ \textbf{1.75}  \\
    \bottomrule
\end{tabular}
\caption{Baseline comparison on BabyAI (top) and AlfWorld (bottom) environments. $\uparrow$ denotes higher is better. $\downarrow$ denotes lower is better. For a fair few-shot comparison, each approach uses 10 in-context examples in BabyAI-Synth and 5 in the remaining environments. The best performing approach is shown in \textbf{Bold}. Error bars are computed using bootstrapped sampling with 10k trials. $(-)$ indicates the metric is not applicable for the approach.}
\label{tab:baseline} \vspace{-1em}
\end{table*}

We evaluate the performance of \approach in AlfWorld \cite{shridhar2020alfworld} and BabyAI \cite{chevalier2018babyai, carta2023grounding} benchmarks, depicted in \cref{fig:domains}. These benchmarks include several features representative of challenges in robot decision-making, thereby making them suitable benchmarks for evaluating \approach. For instance, both benchmarks include a suite of sequential tasks that need to be performed by situated agents. The environments are partially observable to the agent, requiring the agent to explore, navigate and interact with objects to complete tasks described in natural language. Solving these tasks require reasoning over long-horizon in presence of a sparse reward signal, which is challenging for planning, RL and LLM-based decision-making algorithms~\cite{chevalier2018babyai, shridhar2020alfworld, carta2023grounding}.

\textbf{Tasks.} AlfWorld offers a diverse set of household tasks across various difficulty levels. We conduct experiments using the seen and \textsc{unseen} validation sets, which include 140 and 132 task instances, respectively.  The \textsc{seen} set is designed to measure in-distribution generalization, whereas the unseen set measures out-of-distribution generalization. BabyAI, a 2D grid world environment, features 40 levels of varying complexity. We focus on the Synth and BossLevel levels. The \textsc{Synth} level includes single-step instructions, such as ``pick up a ball" or ``go to the red key," while the BossLevel provides more complex, multi-step instructions, such as ``put the yellow ball next to the purple ball, then open the purple door." Each level contains 100 evaluation task instances. 

\textbf{Prompt Design.} We represent robot decisions as Python programs  \cite{singh2023progprompt}. \cref{fig:illustration} illustrates our prompt. The high-level actions are imported as Python functions. Each action is further defined with the types of arguments it requires. Finally, each argument type is defined as a class whose attributes represent the environment objects and their attributes. The interaction at each step is represented by key variables like $\texttt{feasibility\_feedback}$, $\texttt{visible\_objects}$ and $\texttt{inventory}$. We task the LLM to predict the next action as the value of the variable $\texttt{action}$.

\textbf{Language Model.} We use \textsc{GPT-4o} \cite{gpt-4o} as the large language model $LLM$ for generating actions and OpenAI's \textsc{text-embedding-3-large}  \cite{text-embedder} as the embedding model $e$ for encoding instructions into 3072 dimensional vectors. Greedy decoding is applied with a maximum token limit of 200 for the LLM-Planner and 50 for the other approaches. The horizon $(h)$ for high-level actions is set to 30 for AlfWorld, 20 for BabyAI-BossLevel, and 25 for BabyAI-Synth. 

\textbf{Baselines.} We consider the following baselines for comparison, each using language models as high-level planners: (i) ProgPrompt \cite{singh2023progprompt} is a powerful static planner for robotic tasks that generates a complete plan at the start of a task and uses assertion checks to ground the plan to the current state. It is representative of LLM-based agents that do not involve memory or learning from experience.
(ii) LLM-Planner \cite{song2023llm} is a method that employs grounded re-planning, dynamically updating the plan throughout the task. It is a representative approach of more recent LLM-based agents that also utilize retrieval-augmented generation; however, in a different manner than that of \approach.
For each environment, all baselines have access to $100$ training tasks with expert-provided demonstrations. We initialize the memory in \approach using these expert demonstrations. We refer to the initial memory as prior experience, which is updated online based on experience of solving new tasks. 

\textbf{Metrics.} To measure the decision-making performance, we consider three evaluation metrics. (i) Success Rate (SR) measures the fraction of tasks that the planner completed successfully. (ii) Average In-Executability (InExec) is the average number of selected actions that cannot be executed in the environment. (iii) Average Episode Length (Len) is the average number of planning actions that are required to complete a given task. As ProgPrompt is an offline approach, (InExec) and (Len) metrics are not applicable for it.

\section{Results and Discussion}

\begin{table*}
\centering
    \begin{tabular}{lcccccc}
    \\
    \toprule
     &  \multicolumn{3}{c}{BabyAI-Synth}  &  \multicolumn{3}{c}{BabyAI-BossLevel}\\
     \cmidrule(lr){2-4}  \cmidrule(lr){5-7}
     & SR $\uparrow$ & InExec $\downarrow$ & Len $\downarrow$ & SR $\uparrow$  & InExec $\downarrow$ & Len $\downarrow$\\
    \midrule
    \approach & $0.48 \pm 0.1$ & $5.18\pm1.18$ & \textbf{14.82} $\pm$ \textbf{2.14} & \textbf{0.57} $\pm$ \textbf{0.1} & \textbf{3.74} $\pm$ \textbf{0.78}& \textbf{12.48} $\pm$ \textbf{1.49}\\
    \quad with trajectory-level retrieval & \textbf{0.50} $\pm $\textbf{0.1} & $5.30\pm 0.93$ & $15.42\pm 2.04$ & $0.52\pm 0.1$  & $4.22\pm 0.76$ & $13.24\pm 1.43$\\
    \quad without prior experience & $0.44\pm 0.1$ & \textbf{4.67} $\pm$ \textbf{0.88} & $15.92\pm 2.05$ &  $0.54\pm 0.1$ & $4.68\pm 0.88$ & $13.16\pm 1.48$\\ 
    \quad without memory & $0.43\pm 0.1$ & $6.72\pm 1.13$ & $16.23\pm 2.03$ &  $0.37\pm 0.09$ & $6.07\pm 0.99$ & $14.48\pm 1.47$\\
    \bottomrule
    \end{tabular}
\caption{Ablating different components of \approach. $\uparrow$ denotes higher is better. $\downarrow$ denotes lower is better. The best performing approach is shown in \textbf{Bold}. Error bars are computed using bootstrapped sampling with 10k trials. The first row presents the results of the complete \approach framework. The second row corresponds to a variant of \approach with an alternate retrieval function, which retrieves the most similar task trajectory. The third row shows performance of \approach when starting with no prior experience. The last row represents a variant that does not involve a memory component.he tuple $(g^k, c_{-1}^k, f^k, o^k)$.}
\label{tab:ablation} \vspace{-2em}
\end{table*}



\textit{How does \approach compare against other LLM as Agents baselines?}  In \cref{tab:baseline}, we report comparison with the baselines. \approach demonstrates a higher success rate than ProgPrompt across both domains. This can be attributed to ProgPrompt’s lack of memory and critics, which means it does not benefit from interactive learning. By interacting with the environment and retrieving relevant experience, \approach enables more informed decision-making.

\approach also outperforms LLM-Planner in terms of success rate, in-executability, and average episode length. Notably, the success rate improvements range from +0.33 to +0.37 in more challenging environments like BabyAI-BossLevel and AlfWorld-Unseen. Additionally, \approach has lower in-executability (approx. 7 lower in Synth and 16 in Alfworld-Seen), and achieves shorter average episode lengths. While both systems are interactive and utilize retrieval-augmented generation, the key advantage of \approach is the memory of past interactions that includes critics' feedback. By leveraging this memory, \approach can avoid infeasible actions and accomplish tasks with fewer steps. Additionally, the lower episode length achieved by the \approach facilitates a reduction in the overall cost of using LLMs, such as API expenses for closed-source models.

\begin{figure}
    \centering
    \includegraphics[width=0.74\linewidth]{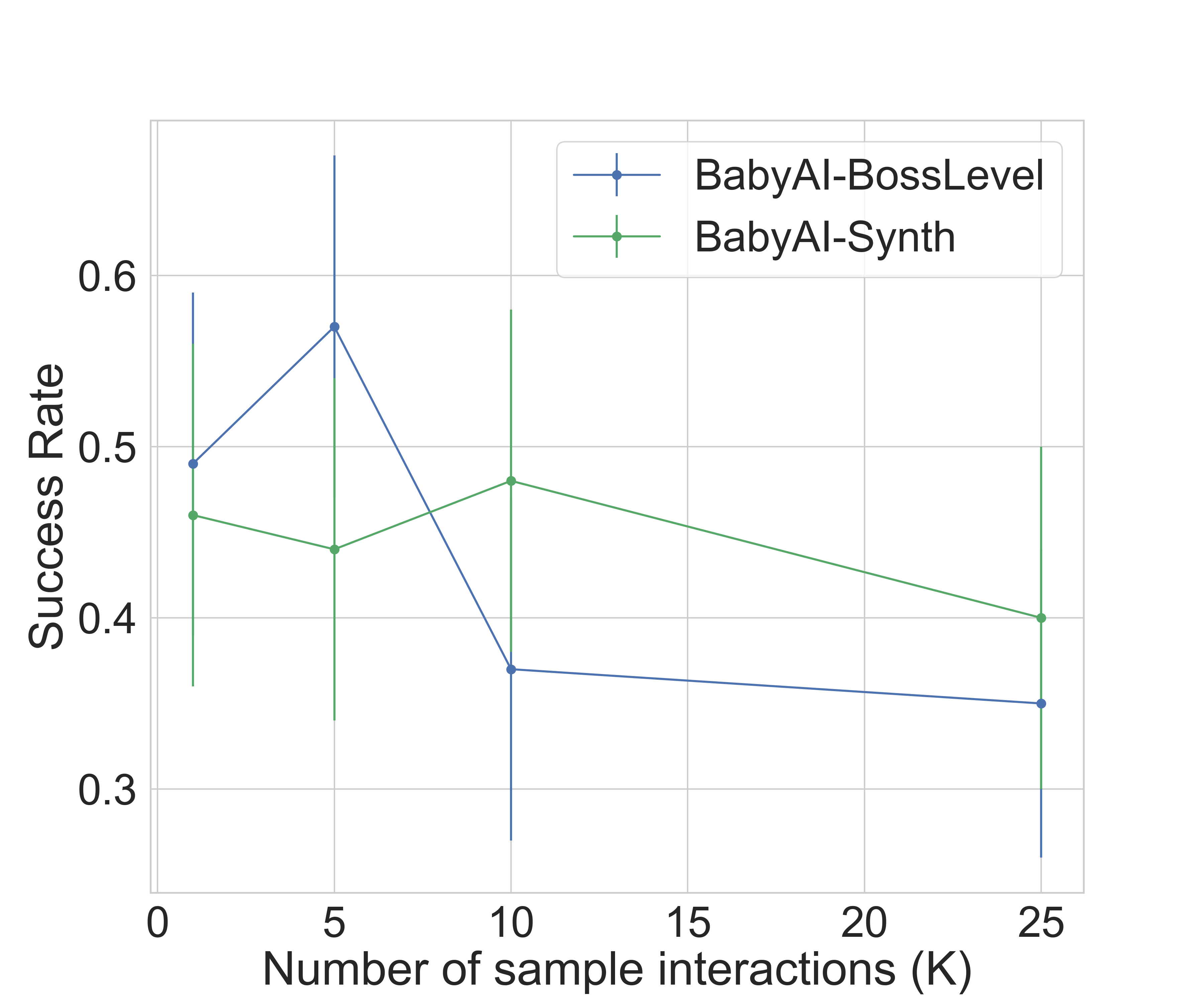}
    \caption{Success Rate as a function of $K$}
    \label{fig:success_rate_over_k} \vspace{-1em}
\end{figure}

\begin{figure}
    \centering
    \includegraphics[width=0.75\linewidth]{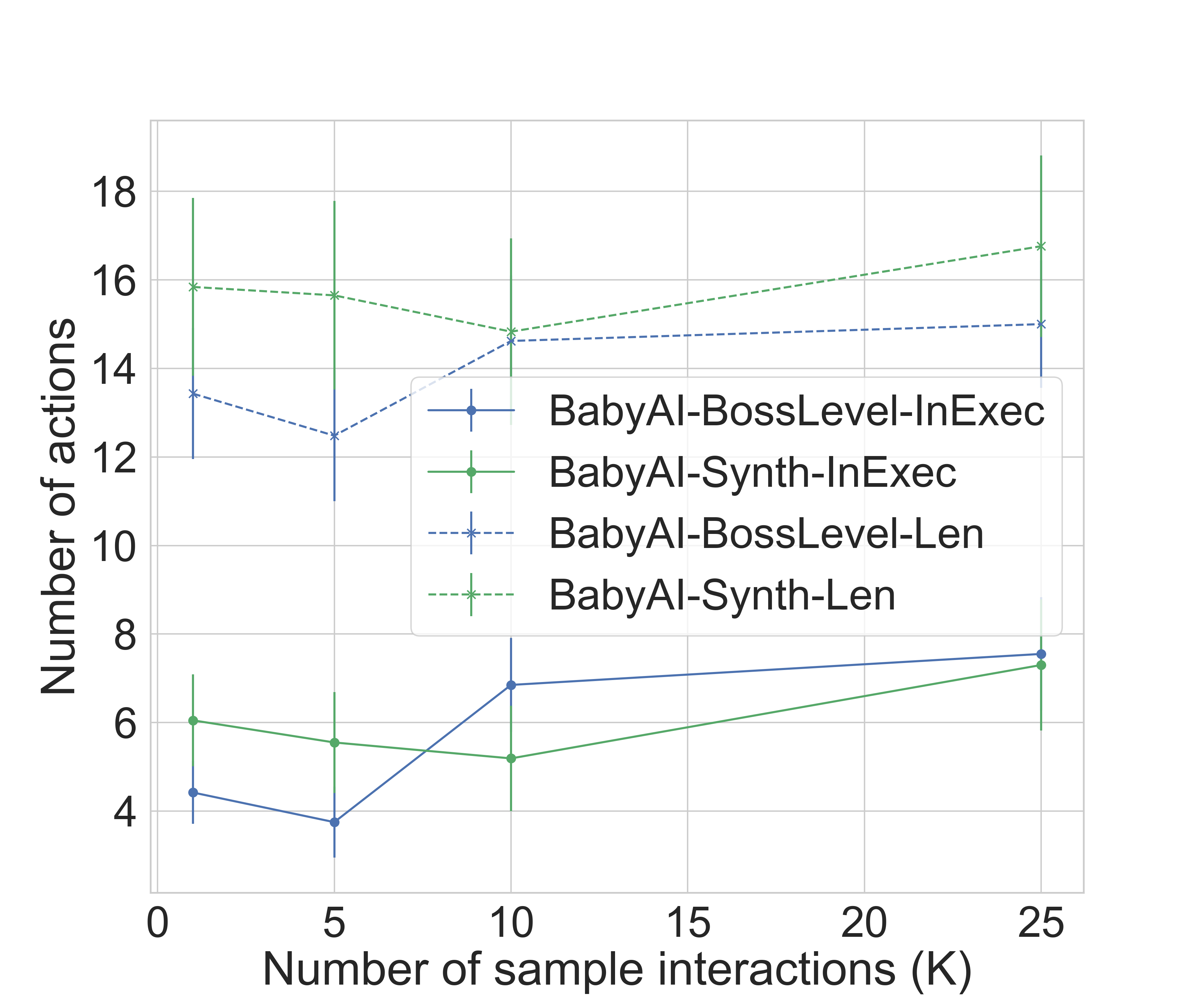}
    \caption{In-Executability and Episode Length as a function of $K$}
    \label{fig:inexec_len_over_k} \vspace{-1em}
\end{figure}

\textit{What is the optimal number of interactions $K$ to use as in-context examples?} We ablate the number of interactions retrieved from memory and evaluate performance on BabyAI environments. The results reported in \cref{fig:success_rate_over_k} show that the success rate improves as $K$ increases, peaking at $K=5$ for BossLevel and $K=10$ for SynthLevel, before beginning to decline. Similarly, in \cref{fig:inexec_len_over_k} we observed that in-executability and average episode length decrease initially but start to rise as $K$ continues to grow. The initial boost in performance can be attributed to the inclusion of more informative interactions,  enhancing the LLM’s decision-making capabilities. The subsequent decline likely stems from the LLM’s sensitivity to irrelevant or noisy context \cite{shi2023large, liu2024lost}. As $K$ increases, the chance of introducing less relevant or low-quality interactions also rises, which can distract the model and degrade its output quality \cite{merth2024superposition}. These trends suggest retrieving a modest number of interactions (between $5$ and $10$) while solving tasks using the \approach framework. 

\textit{How does the choice of retrieval function affect  performance?} 
We examine how retrieving interactions at different levels of granularity impacts performance. Specifically, we compare against an ablation of our approach that utilizes a trajectory-level retrieval function. This ablation first identifies the most relevant task in the memory by computing the cosine similarity between goals, and then extract top $K$ interactions from that task's trajectory. We represent the performance of this variable in the second row of \cref{tab:ablation}. We observe that retrieving at the interaction level generally yields better results, with lower in-executability and shorter episode lengths, while maintaining similar or higher success rates across both BabyAI domains. This suggests that retrieving interactions from a diverse set of tasks provides the language model with richer information than simply retrieving interactions from the most relevant single task.

\textit{How does the presence of memory affect performance?} 
To study the role of memory, we consider a variant of the proposed approach that does not include any interaction memory.
This variant is representative of the LLM-Modulo framework~\cite{kambhampati2024llms}, which includes interaction and critics but no mechanisms for storage or retrieval of experience. 
As reported in the last row of \cref{tab:ablation}, completely removing the memory component leads to a significant drop in performance, with a $0.20$ decrease in success rate on BabyAI-BossLevel and an increase in average episode length by $1.4$ to $2.0$ steps. This demonstrates that storing and retrieving past interactions and feedback significantly improves the decision-making capabilities of the critic-aided language model. 

\textit{How does prior experience affect performance?} 
Lastly, in the third row of \cref{tab:ablation}, we report results of \approach when it is not seeded with any prior expert-generated experience.
Unsurprisingly, we find that prior experience generally helps in sequential decision-making.
Interestingly, even starting our approach with an empty memory (third row, \cref{tab:ablation}) still outperforms the variant that does not include a memory component (fourth row, \cref{tab:ablation}), as the agent can gradually collect experiences of successes and failures, allowing it to learn and improve its decision-making.
\color{black}

\section{Conclusion}
\label{sec:conclusion}
This paper introduces \approach, a framework for solving sequential decision-making tasks by providing LLM-based agents memory of past interactions. Extending the recent LLM-Modulo framework, \approach not only incorporates critic feedback regarding the feasibility of generated actions but also enables agents to remember successes and mistakes and learn from them. \approach demonstrates superior performance on the challenging BabyAI and AlfWorld benchmarks, achieving higher success rates while requiring fewer actions to complete sequential tasks.

In future work, we plan to utilize \approach to solve tasks in other environments involving physical robots, such as FurnitureBench with the Panda robot \cite{heo2023furniturebench}. We also see potential in integrating \approach with existing continual learning frameworks, such as BOSS and Voyager \cite{zhang2023bootstrap, wang2023voyager}, to enable learning from experience at multiple layers of abstractions: namely, skills and interactions. Another avenue is to explore tunable retrieval models that can anticipate future needs to further enhance the agent's performance \cite{weijia2023replug, jiang2023active}. Finally, 
we are interested in studying how \approach can enhance end-user programming of complex robot behaviors by leveraging user commands, experience and critiques.


\bibliographystyle{IEEEtran}  
\bibliography{root}
\end{document}